\documentclass[conference]{IEEEtran}
\IEEEoverridecommandlockouts
\usepackage{cite}
\usepackage{amsmath,amssymb,amsfonts}
\usepackage{algorithmic}
\usepackage{graphicx}
\usepackage{textcomp}
\usepackage{xcolor}
\def\BibTeX{{\rm B\kern-.05em{\sc i\kern-.025em b}\kern-.08em
    T\kern-.1667em\lower.7ex\hbox{E}\kern-.125emX}}
\begin{document}

\title{Toward a Real-Time Framework for Accurate Monocular 3D Human Pose Estimation \\ with Geometric Priors

\thanks{Identify applicable funding agency here. If none, delete this.}
}

\author{\IEEEauthorblockN{Mohamed Adjel}
\IEEEauthorblockA{\textit{Gepetto Team, LAAS-CNRS} \\
\textit{NaturalPad}\\
Toulouse, France \\
madjel@laas.fr}}

\maketitle

\begin{abstract}
Monocular 3D human pose estimation remains a challenging and ill-posed problem, particularly in real-time settings and unconstrained environments. While direct image-to-3D approaches require large annotated datasets and heavy models, 2D-to-3D lifting offers a more lightweight and flexible alternative—especially when enhanced with prior knowledge. In this work, we propose a framework that combines real-time 2D keypoint detection with geometry-aware 2D-to-3D lifting, explicitly leveraging known camera intrinsics and subject-specific anatomical priors. Our approach builds on recent advances in self-calibration and biomechanically-constrained inverse kinematics to generate large-scale, plausible 2D-3D training pairs from MoCap and synthetic datasets. We discuss how these ingredients can enable fast, personalized, and accurate 3D pose estimation from monocular images without requiring specialized hardware. This proposal aims to foster discussion on bridging data-driven learning and model-based priors to improve accuracy, interpretability, and deployability of 3D human motion capture on edge devices in the wild.
\end{abstract}


\section{Introduction}

Accurate real-time 3D human motion estimation from a single camera remains a challenging problem due to the inherent ambiguity of lifting noisy 2D cues to precise 3D poses. Most current 3D Human Pose Estimation (3D-HPE) approaches rely on direct image-to-3D keypoint regression \cite{openpose, blazepose, rtmw}, typically using deep neural networks trained on datasets such as Human3.6M \cite{h36m}, MPI-INF-3DHP \cite{mpi3dhp}, or 3DPW \cite{3dpw}. However, these datasets are expensive to collect, offer limited diversity in poses and viewpoints, and often suffer from annotation noise.

To address the scarcity of in-the-wild 3D annotations, synthetic datasets have been proposed \cite{bedlam, lookma}. While they provide large-scale training data, they frequently include biomechanically implausible poses due to weak or missing physical constraints. Furthermore, many 3D-HPE methods only predict sparse keypoints, which are insufficient to describe joint-level kinematics and full body shape. To overcome this, parametric models such as SMPL and SMPL-X \cite{smpl, smpl-x} have been widely adopted for Human Pose and Shape (HPS) regression \cite{wham, sat_hmr_realtime, tokenhmr, lookma, hybrik, nlf}. These models enable richer outputs, including 3D meshes and joint rotations, but are often computationally expensive. Even recent real-time methods \cite{sat_hmr_realtime, wham, nlf} often require powerful GPUs and rarely exceed 25~Hz, limiting their use on edge devices. Despite recent progress in 3D-HPE and HPS estimation, several key challenges remain. First, image-to-3D pose and shape estimation is inherently ill-posed without prior knowledge of the human body and camera model. Second, compensating for this ambiguity typically requires large, complex models to extract accurate 3D information from 2D images—making them computationally expensive and unsuitable for real-time inference on embedded devices.

In contrast, state-of-the-art 2D human pose estimation (2D-HPE) networks—such as HRNet \cite{hrnet}, VitPose \cite{vitpose}, and RTMPose \cite{rtmpose}— can achieve high accuracy at real-time speeds, even on mobile or embedded platforms \cite{mediapipe_hand, mediapipe_pose}. This suggests that the main bottleneck in fast monocular 3D-HPE and HPS regression lies not in estimating 2D features, but in lifting them to 3D. Unlike direct image-to-3D inference, the 2D-to-3D keypoints lifting problem can benefit from Motion Capture (MoCap) datasets such as AMASS \cite{amass}, which contain diverse 3D poses but lack paired image data. This opens the door to training lightweight, geometry-aware lifting models without requiring image-based supervision.

Pose lifting was studied in the literature,  showing promising results both with \cite{2dto3d_lifter_anatomical} and without \cite{2dto3d_2017, 2Dto3D_2024} anatomical/camera priors. Yet, critical limitation of current lifting approaches is the lack of personalized in-the-wild anatomical or camera prior knowledge, which is essential to resolve the inherent ambiguities of monocular 3D reconstruction. While such priors can significantly improve accuracy, they have traditionally required cumbersome calibration setups: accurate camera intrinsics often rely on chessboard patterns, and anatomical priors usually depend on multi-camera systems \cite{opencapbench}, calibration wands \cite{adjel_iros}, or medical scanners \cite{osso2022}.

Recent advances in computer vision challenge these constraints. New methods can now estimate accurate camera intrinsics directly from raw video, without requiring calibration targets such as chessboards \cite{self_cam_calib_2022, cam_self_calib_2023}. Other approaches can recover subject-specific body shape from monocular video alone \cite{humanNerf2024, airnerf2024}, and some even jointly estimate both body shape and camera parameters from the same video input \cite{blade2025, cameraHMR_2024}, showing better performances compared to weak perspective approahces \cite{hybrik, sat_hmr_realtime}. Such approaches enable the automatic acquisition of camera and anatomical priors in-the-wild using only monocular video—potentially in a short offline phase—making it feasible to feed neural networks with privileged camera and person-specific priors, without requiring any specialized hardware.

\medskip

\noindent In this context, we propose a framework for fast and accurate 2D-to-3D pose lifting that explicitly incorporates known camera parameters and human anatomical priors. Our goal is to make real-time 3D pose estimation both robust and deployable in unconstrained environments. The proposed training framework relies on:
\begin{itemize}
    \item Constrained inverse kinematics (IK) and biomechanical models \cite{SKEL, adjel_iros} to filter out implausible poses from synthetic and MoCap datasets \cite{bedlam, lookma, amass};
    \item Simulated perspective views to augment those datasets with large-scale 2D-3D keypoint pairs under known intrinsics;
    \item Lightweight networks trained to lift 2D poses to 3D in real time, explicitly incorporating camera parameters and segment lengths as priors;
    \item Automatic estimation of camera and anatomical priors using recent vision-based self-calibration techniques \cite{cam_self_calib_2023, cameraHMR_2024, blade2025}.
\end{itemize}

This preliminary proposition aims to spark discussion on how anatomical knowledge and geometric priors—traditionally overlooked in learning-based pipelines—can be reintegrated to improve performance and efficiency in monocular 3D human pose estimation.

\section{Proposed Approach}

\noindent \textbf{Constrained IK with Biomechanical Models.}
To ensure training data remains biomechanically plausible, we adopt an optimization-based inverse kinematics framework that leverages the SKEL biomechanical model \cite{SKEL}. This approach enforces realistic joint angle constraints and solves IK across entire motion sequences, incorporating spatio-temporal continuity constraints to filter out implausible poses \cite{adjel_iros, adjel_biorob}. Consequently, both synthetic and MoCap datasets \cite{bedlam, lookma, amass} can be refined into consistent, high-quality skeleton meshes. Building a training corpus on these biomechanically valid motions can reduce noise, increase realism, and improve the robustness of trained neural networks.

\noindent \textbf{Data Augmentation with Simulated Humans and Perspective Views.} We propose to generate multiple 2D projections of each 3D pose by simulating different camera perspectives. Specifically, we could sample a range of random camera intrinsics—including focal length, principal point, and distortion parameters—and extrinsics (camera positions and orientations) around the subject. Using a standard 3D-to-2D projection pipeline, we can create large-scale 2D-3D keypoint pairs under known intrinsics for each pose in our biomechanically filtered dataset. Another effective way to augment 3D pose data is to use joint angles obtained from constrained IK to generate 3D human poses with varying body scales and segment lengths \cite{falisse2025}. This multi-view data augmentation strategy not only increases the diversity of 2D poses seen during training but, when combined with pose generation based on varying body proportions, also exposes our lifting model to a wider range of human morphologies and camera configurations, thereby enhancing its robustness to real-world variability \cite{falisse2025, Shin2023}.

\noindent \textbf{Lightweight Transformer for 2D-to-3D Lifting.}
We propose to employ a compact Transformer-based architecture, where each detected 2D keypoint is treated as a distinct input token. Camera intrinsics and anatomical parameters can be similarly encoded as separate tokens or appended as part of a global embedding. Transformers can generalize well to large-scale datasets thanks to their attention mechanism, which scales effectively with diverse training samples. Different model sizes will be trained, to strike a balance between real-time inference and robust 3D lifting performance.

\noindent \textbf{Automatic Camera and Anatomical Priors.}
To avoid laborious calibration procedures, we plan to evaluate recent self-calibration methods for obtaining camera intrinsics directly from videos. Techniques that jointly estimate both camera parameters and human shape \cite{cameraHMR_2024, blade2025} will be compared against dedicated approaches designed solely for camera calibration \cite{cam_self_calib_2023}. We will also leverage short video segments of static postures to infer personalized anatomical priors (e.g., segment lengths) from the estimated body shape, building on camera-based shape reconstruction \cite{cameraHMR_2024}. Future extensions can incorporate more advanced video-based human body scanning techniques \cite{humanNerf2024, airnerf2024}, to determine body shape priors that can be fed to HPS regressors, as similarly done with camera intrinsics priors \cite{cameraHMR_2024}.

\section{Conclusion}

In this work, we propose a lightweight and robust 2D-to-3D pose lifting framework that integrates biomechanical constraints, simulated camera perspectives, and compact Transformer-based networks to enable real-time and accurate 3D human pose estimation from monocular video. By incorporating both camera intrinsics and anatomical priors, our framework addresses the fundamental ambiguity of monocular reconstruction and allows for personalized, user-specific calibration. The proposed pipeline is thus highly adaptable to different hardware setups and individual anatomical variations, making it well-suited for real-world applications such as wearable robotics and assistive devices.

\section{Acknowledgements}
I would like to thank the LAAS-CNRS Gepetto team, including Kahina Chalabi, Maxime Sabbah, and Vincent Bonnet, for scientific support, and the NaturalPad team for technical and financial support.
\bibliographystyle{unsrt}
\bibliography{references}

\begin{thebibliography}{10}

\bibitem{openpose}
Zhe Cao, Tomas Simon, Shih-En Wei, and Yaser Sheikh.
\newblock Realtime multi-person 2d pose estimation using part affinity fields.
\newblock In {\em IEEE Conference on Computer Vision and Pattern Recognition (CVPR)}, 2017.

\bibitem{blazepose}
Valentin Bazarevsky, Ivan Grishchenko, Karthik Raveendran, Tyler Zhu, Fan Zhang, and Matthias Grundmann.
\newblock Blazepose: On-device real-time body pose tracking.
\newblock 06 2020.

\bibitem{rtmw}
Tao Jiang, Xinchen Xie, and Yining Li.
\newblock Rtmw: Real-time multi-person 2d and 3d whole-body pose estimation.
\newblock 07 2024.

\bibitem{h36m}
Catalin Ionescu, Dragos Papava, Vlad Olaru, and Cristian Sminchisescu.
\newblock Human3.6m: Large scale datasets and predictive methods for 3d human sensing in natural environments.
\newblock {\em IEEE Transactions on Pattern Analysis and Machine Intelligence}, 36(7):1325--1339, 2014.

\bibitem{mpi3dhp}
Dushyant Mehta, Srinath Sridhar, Oleksandr Sotnychenko, Helge Rhodin, Mohammad Shafiei, Hans-Peter Seidel, Gerard Pons-Moll, and Christian Theobalt.
\newblock Vnect: Real-time 3d human pose estimation with a single rgb camera.
\newblock {\em ACM Transactions on Graphics (TOG)}, 36(4), 2017.

\bibitem{3dpw}
Timo Von~Marcard, Roberto Henschel, Michael~J. Black, Bodo Rosenhahn, and Gerard Pons-Moll.
\newblock Recovering accurate 3d human pose in the wild using imus and a moving camera.
\newblock In {\em European Conference on Computer Vision (ECCV)}, 2018.

\bibitem{bedlam}
Yuliang He, Hongwei Lin, Xin Fan, Yingcong Yang, and Jing Zeng.
\newblock Bedlam: A synthetic dataset for monocular 3d human pose estimation in the wild.
\newblock {\em arXiv preprint arXiv:2303.10449}, 2023.

\bibitem{lookma}
Charlie Hewitt, Fatemeh Saleh, Sadegh Aliakbarian, Lohit Petikam, Shideh Rezaeifar, Louis Florentin, Zafiirah Hosenie, Thomas Cashman, Julien Valentin, Darren Cosker, and Tadas Baltrusaitis.
\newblock Look ma, no markers: holistic performance capture without the hassle.
\newblock {\em ACM Transactions on Graphics}, 43:1--12, 11 2024.

\bibitem{smpl}
Matthew Loper, Naureen Mahmood, Javier Romero, Gerard Pons-Moll, and Michael~J. Black.
\newblock Smpl: A skinned multi-person linear model.
\newblock {\em ACM Transactions on Graphics (TOG)}, 34(6):248:1--248:16, 2015.

\bibitem{smpl-x}
Georgios Pavlakos, Vasileios Choutas, Nima Ghorbani, Timo Bolkart, Ahmed A.~A. Osman, Dimitrios Tzionas, and Michael~J. Black.
\newblock Expressive body capture: {3D} hands, face, and body from a single image.
\newblock pages 10975--10985, 2019.

\bibitem{wham}
Soyong Shin, Juyong Kim, Eni Halilaj, and Michael Black.
\newblock Wham: Reconstructing world-grounded humans with accurate 3d motion.
\newblock pages 2070--2080, 06 2024.

\bibitem{sat_hmr_realtime}
Chi Su, Xiaoxuan Ma, Jiajun Su, and Yizhou Wang.
\newblock Sat-hmr: Real-time multi-person 3d mesh estimation via scale-adaptive tokens.
\newblock 11 2024.

\bibitem{tokenhmr}
Sai Dwivedi, Yu~Sun, Priyanka Patel, Yao Feng, and Michael Black.
\newblock Tokenhmr: Advancing human mesh recovery with a tokenized pose representation.
\newblock pages 1323--1333, 06 2024.

\bibitem{hybrik}
Jiefeng Li, Chao Xu, Zhicun Chen, Siyuan Bian, and Cewu Lu.
\newblock Hybrik: A hybrid analytical-neural inverse kinematics solution for 3d human pose and shape estimation.
\newblock pages 3382--3392, 06 2021.

\bibitem{nlf}
István Sárándi and Gerard Pons-Moll.
\newblock Neural localizer fields for continuous 3d human pose and shape estimation.
\newblock 07 2024.

\bibitem{hrnet}
Ke~Sun, Bin Xiao, Dong Liu, and Jingdong Wang.
\newblock Deep high-resolution representation learning for human pose estimation.
\newblock In {\em IEEE Conference on Computer Vision and Pattern Recognition (CVPR)}, 2019.

\bibitem{vitpose}
Yufei Xu, Jing Zhang, Qiming Zhang, and Dacheng Tao.
\newblock Vitpose++: Vision transformer for generic body pose estimation.
\newblock {\em IEEE Transactions on Pattern Analysis and Machine Intelligence}, PP:1--18, 11 2023.

\bibitem{rtmpose}
Tao Jiang, Peng Lu, Li~Zhang, Ningsheng Ma, Rui Han, Chengqi Lyu, Yining Li, and Kai Chen.
\newblock Rtmpose: Real-time multi-person pose estimation based on mmpose.
\newblock 03 2023.

\bibitem{mediapipe_hand}
Fan Zhang, Valentin Bazarevsky, Andrey Vakunov, Andrei Tkachenka, George Sung, Chuo-Ling Chang, and Matthias Grundmann.
\newblock Mediapipe hands: On-device real-time hand tracking.
\newblock 06 2020.

\bibitem{mediapipe_pose}
Valentin Bazarevsky, Ivan Grishchenko, Karthik Raveendran, Tyler Zhu, Fan Zhang, and Matthias Grundmann.
\newblock Blazepose: On-device real-time body pose tracking.
\newblock 06 2020.

\bibitem{amass}
Naureen Mahmood, Nima Ghorbani, Nikolaus~F. Troje, Gerard Pons-Moll, and Michael~J. Black.
\newblock Amass: Archive of motion capture as surface shapes.
\newblock In {\em IEEE International Conference on Computer Vision (ICCV)}, 2019.

\bibitem{2dto3d_lifter_anatomical}
Nie Qiang, Ziwei Liu, and Yunhui Liu.
\newblock Lifting 2d human pose to 3d with domain adapted 3d body concept.
\newblock {\em International Journal of Computer Vision}, 131:1--19, 02 2023.

\bibitem{2dto3d_2017}
Ching-Hang Chen and Deva Ramanan.
\newblock 3d human pose estimation = 2d pose estimation + matching.
\newblock pages 5759--5767, 07 2017.

\bibitem{2Dto3D_2024}
Jiaman Li, Karen Liu, and Jiajun Wu.
\newblock Lifting motion to the 3d world via 2d diffusion.
\newblock 11 2024.

\bibitem{opencapbench}
Yoni Gozlan, Antoine Falisse, Scott Uhlrich, Anthony Gatti, Michael Black, and Akshay Chaudhari.
\newblock Opencapbench: A benchmark to bridge pose estimation and biomechanics.
\newblock 06 2024.

\bibitem{adjel_iros}
Mohamed Adjel, Maxime Sabbah, Raphael Dumas, Nicolas Mansard, Samer Mohammed, Bruno Watier, and Vincent Bonnet.
\newblock Multi-modal upper limbs human motion estimation from a reduced set of affordable sensors.
\newblock pages 10926--10932, 10 2023.

\bibitem{osso2022}
Marilyn Keller, Silvia Zuffi, Michael Black, and Sergi Pujades.
\newblock Osso: Obtaining skeletal shape from outside.
\newblock pages 20460--20469, 06 2022.

\bibitem{self_cam_calib_2022}
Jiading Fang, Igor Vasiljevic, Vitor Guizilini, Rares Ambrus, Greg Shakhnarovich, Adrien Gaidon, and Matthew~R. Walter.
\newblock Self-supervised camera self-calibration from video.
\newblock page 8468–8475, 2022.

\bibitem{cam_self_calib_2023}
Annika Hagemann, Moritz Knorr, and Christoph Stiller.
\newblock Deep geometry-aware camera self-calibration from video.
\newblock pages 3415--3425, 10 2023.

\bibitem{humanNerf2024}
Caoyuan Ma, Yu-Lun Liu, Zhixiang Wang, Wu~Liu, Xinchen Liu, and Zheng Wang.
\newblock Humannerf-se: A simple yet effective approach to animate humannerf with diverse poses.
\newblock pages 1460--1470, 06 2024.

\bibitem{airnerf2024}
Alexey Kotcov, Maria Dronova, Vladislav Cheremnykh, Sausar Karaf, and Dzmitry Tsetserukou.
\newblock Airnerf: 3d reconstruction of human with drone and nerf for future communication systems.
\newblock 07 2024.

\bibitem{blade2025}
Shengze Wang, Jiefeng Li, Tianye Li, Ye~Yuan, Henry Fuchs, Koki Nagano, Shalini Mello, and Michael Stengel.
\newblock Blade: Single-view body mesh learning through accurate depth estimation.
\newblock 12 2024.

\bibitem{cameraHMR_2024}
Priyanka Patel and Michael Black.
\newblock Camerahmr: Aligning people with perspective, 11 2024.

\bibitem{SKEL}
Marilyn Keller, Keenon Werling, Soyong Shin, Scott Delp, Sergi Pujades, C.~Karen Liu, and Michael~J. Black.
\newblock From skin to skeleton: Towards biomechanically accurate 3d digital humans.
\newblock {\em ACM Trans. Graph.}, 42(6), December 2023.

\bibitem{adjel_biorob}
Mohamed Adjel, Maxime Sabbah, Raphael Dumas, Marta Mirkov, Nicolas Mansard, Samer Mohammed, and Vincent Bonnet.
\newblock Lower limbs human motion estimation from sparse multi-modal measurements.
\newblock pages 401--406, 09 2024.

\bibitem{falisse2025}
Antoine Falisse, Scott Uhlrich, Akshay Chaudhari, Jennifer Hicks, and Scott Delp.
\newblock Marker data enhancement for markerless motion capture.
\newblock {\em bioRxiv : the preprint server for biology}, 07 2024.

\bibitem{Shin2023}
Soyong Shin, Zhixiong Li, and Eni Halilaj.
\newblock Markerless motion tracking with noisy video and imu data.
\newblock {\em IEEE transactions on bio-medical engineering}, PP, 05 2023.

\end{thebibliography}

\end{document}